%% file: fedseq.tex
\documentclass[a4paper,conference]{IEEEtran}

\input{tools}

\usepackage{multirow}
\usepackage{amssymb}
\usepackage{amsfonts} 
\usepackage{nicefrac}
\usepackage{booktabs}
\usepackage{hyperref}

\begin{document}
%
\title{Speeding up Heterogeneous Federated Learning with Sequentially Trained Superclients}

\author{\IEEEauthorblockN{Riccardo Zaccone\textsuperscript{*}, Andrea Rizzardi\textsuperscript{*}, Debora Caldarola, Marco Ciccone, Barbara Caputo}
\IEEEauthorblockA{Politecnico di Torino, Turin, Italy\\
\textsuperscript{*}Equal contributors and corresponding authors: \{name.surname\}@studenti.polito.it}
}


%


\maketitle

\begin{abstract}
\input{sections/0.abstract}
\end{abstract}


%
\IEEEpeerreviewmaketitle

\input{sections/1.introduction}
\input{sections/2.related_works}
\input{sections/3.method}
\input{sections/4.experiments}
\input{sections/5.conclusion}






%

\bibliographystyle{plain}
\bibliography{biblio}

\newpage
\onecolumn

\appendix

\input{supplementary}

\end{document}

%% file: tools.tex
\usepackage{cite}
\ifCLASSINFOpdf
  \usepackage[pdftex]{graphicx}
\else
  \usepackage[dvips]{graphicx}
\fi
\usepackage{amsmath}
\usepackage{dsfont}
\usepackage{multirow}
\usepackage{arydshln}
\usepackage{algorithm2e}
\DontPrintSemicolon 
\RestyleAlgo{ruled}
\usepackage{xfrac}
\usepackage[usenames,dvipsnames]{xcolor}
\usepackage{soul}
\usepackage{algorithmic}
\usepackage{fixltx2e}
\usepackage{stfloats, subfig}
\usepackage{url}
\usepackage{multirow}
\usepackage{amssymb}
\usepackage{amsfonts} 
\usepackage{nicefrac}
\usepackage{booktabs}
\usepackage{xpunctuate}
\usepackage{xcolor-material}
\usepackage{color, colortbl}
\usepackage{soul} 
\usepackage{xspace}
\usepackage{amsmath}
\usepackage{hyphenat}
\usepackage{arydshln}

\newcommand{\quotes}[1]{``#1''} 

\makeatletter
\DeclareRobustCommand\onedot{\futurelet\@let@token\@onedot}
\def\@onedot{\ifx\@let@token.\else.\null\fi\xspace}

\def\ie{\emph{i.e}\onedot}

\def\etal{\emph{et al}\onedot}
\makeatother

\makeatletter
\let\origparagraph\paragraph
\renewcommand\paragraph{\@ifstar{\starparagraph}{\nostarparagraph}}
\newcommand\nostarparagraph[1]
{\origparagraph{#1}\paragraphpostlude}
\newcommand\starparagraph[1]
{\paragraphprelude\origparagraph*{#1}\paragraphpostlude}
\newcommand\paragraphprelude{%
  \vspace{-14pt}%
}
\newcommand\paragraphpostlude{%
}

\makeatletter
 \def\SOUL@hlpreamble{%
 \setul{}{3.2ex}
 \let\SOUL@stcolor\SOUL@hlcolor
 \SOUL@stpreamble
 }
\makeatother

\DeclareMathOperator*{\argmin}{arg\,min}

\newcommand{\hlc}[2][yellow]{{%
    \colorlet{foo}{#1}%
    \sethlcolor{foo}\hl{#2}}%
}

\DeclareRobustCommand{\methodShort}
{FedSeq}

%% file: sections/0.abstract.tex
Federated Learning (FL) 
allows training machine learning models in privacy-constrained scenarios by enabling the cooperation of edge devices without requiring local data sharing.
This approach raises several challenges due to the different statistical distribution of the local datasets and the clients' computational heterogeneity.
In particular, the presence of highly non-i.i.d. data severely impairs both the performance of the trained neural network and its convergence rate, increasing the number of communication rounds requested to reach a performance comparable to that of the centralized scenario.
As a solution, we propose FedSeq, a novel framework
leveraging the sequential training of subgroups of heterogeneous clients, \ie \textit{superclients}, to emulate the centralized paradigm in a privacy-compliant way. Given a fixed budget of communication rounds,
we show that \methodShort~outperforms or match several state-of-the-art federated algorithms in terms of final performance and speed of convergence. Finally, our method can be easily integrated with other approaches available in the literature. Empirical results show that combining existing algorithms with FedSeq further improves its final performance and convergence speed. We test our method on CIFAR-10 and CIFAR-100 and prove its effectiveness in both i.i.d. and non-i.i.d. scenarios.\footnote{Code available at
\texttt{\url{https://github.com/RickZack/FedSeq}}.}



%% file: sections/1.introduction.tex
\section{Introduction}
In 2017, McMahan \etal \cite{mcmahan2017communication} introduced Federated Learning (FL) to train machine learning models in a distributed fashion while respecting privacy constraints on the edge devices.
In FL, the clients are involved in an iterative two-step process over several communication rounds: (i) independent training on edge devices on local datasets, and (ii) aggregation of the updated models into a shared global one on the server-side. This approach is usually effective in homogeneous scenarios, but fails to reach comparable performance against non-i.i.d. data. In particular, 
it has been shown that the non-iidness of local datasets leads to unstable and slow convergence \cite{li2020federated}, suboptimal performance \cite{zhao2018federated, li2021NonIIDSilos} or even model divergence \cite{mcmahan2017communication}.
 \begin{figure}[t]
        \centering
        \includegraphics[width=1\linewidth]{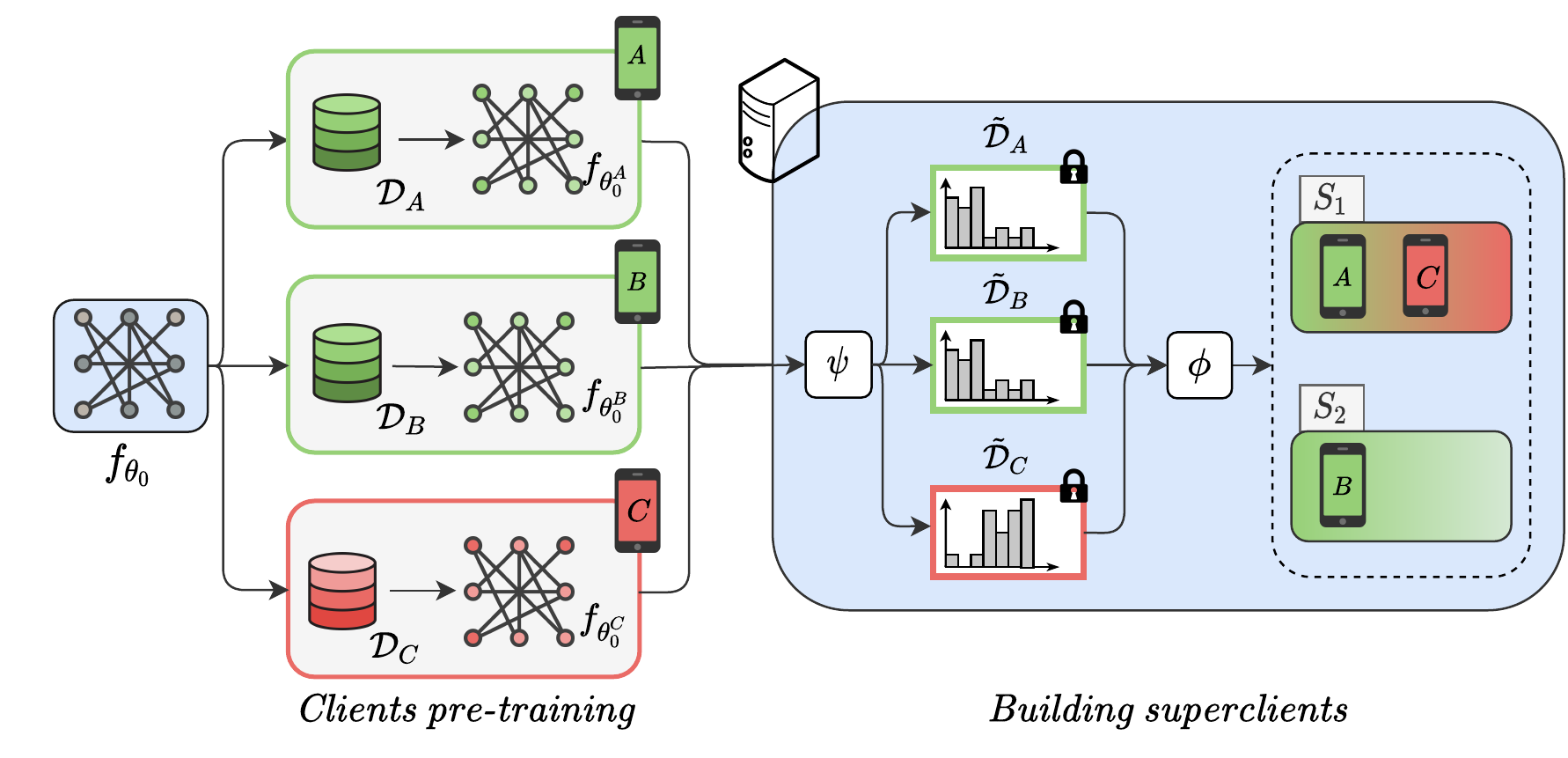}
        \caption{Building superclients with FedSeq. i) The initial model $f_{\theta_0}$ is sent to the clients, where is trained to fit the local distributions $\mathcal{D}_k$. ii) On the server-side, according to an approximator $\psi$, the trained models $f_{\theta_0^k}$ are used to estimate the clients' distributions $\Tilde{\mathcal{D}}_k$. $\phi$ builds the superclients, grouping together clients having different distributions (A and C), while dividing similar ones (A and B).}
        \label{fig:fedseq_teaser}
    \end{figure}
Several lines of research emerged to address the statistical challenges of FL: client drift mitigation aims at regularizing the local objective in order to make it closer to the global one \cite{li2020federated, karimireddy2020scaffold, acar2021federated}; multi-task approaches treat each distribution as a \textit{task} and focus on fitting separate but related models simultaneously \cite{smith2017federated}; FCL integrates Continual Learning (CL) in the FL setting by allowing each client to have a privately accessible sequence of tasks \cite{usmanova2021distillation}; data sharing approaches use small amounts of public or synthesized i.i.d. data to help build a more balanced data distribution \cite{zhao2018federated}.

In this work, we tackle the problems of i) \textit{non identical class distribution}, meaning that for a given pair instance-label $(x,y) \sim P_k(x,y)$, $P_k(y)$ varies across edge devices $k$ while $P(y|x)$ is identical, and ii) \textit{small local dataset cardinality}. 
Inspired by the differences with the standard centralized training procedure, which bounds any FL algorithm, we introduce Federated Learning via Sequential Superclients Training (\methodShort), a novel algorithm that leverages sequential training among subgroups of clients to tackle statistical heterogeneity. 
We simulate the presence of homogeneous and larger datasets without violating the privacy constraints: clients having different distributions are grouped, forming a \textit{superclient} based on a dissimilarity metric. 
Then, within each superclient, the global model is trained sequentially, and the updates are finally combined on the server-side. Intuitively, this scheme resembles the training on devices having larger and less unbalanced datasets, falling into a favorable scenario for FL.
To the best of our knowledge, this is the first federated algorithm to employ such a sequential training on clients grouped by their dissimilarity.
To summarize, our main contributions are:
\begin{itemize}
    \item We introduce \methodShort, a new federated algorithm which learns from groups of sequentially-trained clients, namely \textit{superclients}.
    \item We introduce two lightweight procedures to estimate the probability distribution of a client and analyze how they affect the ability of grouping algorithms to produce better superclients. We evaluate two strategies, comparing them with the naïve random assignment, showing the impact of groups quality on the algorithm convergence.
    \item We show that our method outperforms the state-of-the-art in terms of convergence performance and speed in both i.i.d. and non-i.i.d. scenarios 
   
\end{itemize}

%% file: sections/2.related_works.tex
\section{Related works}

Recent years have seen a growing interest in Federated Learning \cite{kairouz2019advances,li2020federated_survey,xu2021federated}. In realistic federated scenarios, a major challenge is posed by the non-i.i.d. and highly unbalanced distribution of the clients' data, also known as \textit{statistical heterogeneity} \cite{li2019convergence,hsu2019measuring}.

FedAvg \cite{mcmahan2017communication} defines the standard optimization method in FL, where a global model is obtained as a weighted average of local models trained on clients' private data. However, in heterogeneous settings, the local optimization objectives drift from each other, leading to different local models which are hard to be aggregated~\cite{karimireddy2020scaffold}. Several works demonstrate how the convergence rates of FedAvg get worse with the increase of clients heterogeneity~\cite{wang2019adaptive, li2020federated, khaled2019first, li2019convergence, hsu2020federated}. SCAFFOLD~\cite{karimireddy2020scaffold} tries to mitigate this issue by introducing control variates, while FedProx~\cite{li2020federated} adds a proximal term to the local loss function. FedDyn~\cite{acar2021federated} dynamically updates the local objective to ensure the asymptotic alignment of the global and devices solutions.
Server-side optimizers~\cite{reddi2020adaptive,hsu2020federated} have been also introduced for coping with FedAvg lack of adaptivity. In \cite{michieli2021all}, it is showed how fair model aggregation is beneficial when clients observe non-i.i.d. data. While referring mainly to~\cite{mcmahan2017communication} for the aggregation scheme, our work revises the standard framework to account for statistical heterogeneity.



As the learned local model under-represents the deducible patterns from the missing classes, \cite{zhao2018federated} shows how sharing a small set of public data among the clients leads to notable improvements. A similar approach is followed by \cite{li2019fedmd}, where the public data enables knowledge distillation. Similarly to \cite{li2019fedmd}, we keep the public data on the server-side, with the different purpose of using them to estimate the clients' data distribution in a privacy-compliant way. Unlike \cite{li2019fedmd,zhao2018federated}, such data is never used at training time.

Another line of work tackles the problem from a multitask perspective \cite{caruana1997multitask}, where each client is treated as a different task \cite{smith2017federated,fallah2020personalized}. In \cite{sattler2020clustered,xie2021multi,briggs2020federated,caldarola2021cluster},
\cite{kopparapu2020fedfmc} clients with similar tasks are clustered together and a specialized model is assigned to each cluster. In \cite{caldarola2021cluster}, tasks are identified using a domain classifier learned via knowledge distillation and then addressed by the means of a graph, while in our method, following the same approach of \cite{sattler2020clustered,xie2021multi,briggs2020federated}, the locally trained model are used to approximate the clients' data distribution.
Unlike those works, \methodShort~exploits clustering methods to group together clients having distant distributions, in order to obtain an underlying homogeneous dataset within each group, \ie \textit{superclient}. Our approach also relates to the \quotes{\textit{anti-clustering}} literature~\cite{papenberg2021using, anticlustering_analysis}, where the goal is to build similar groups from dissimilar elements~\cite{set_partition}.
From here on we will refer to such techniques as \quotes{\textit{grouping algorithms}}.
Finally, \methodShort~also relates to peer-to-peer (P2P) methods for FL~\cite{roy2019braintorrent,hu2019decentralized} by sharing models between clients belonging to the same superclient. Unlike such works, we keep the central server as a proxy between clients and prioritize FL's statistical challenges rather than communication costs.



%% file: sections/3.method.tex
\section{Method}
\label{sec:method}
\subsection{Problem formulation}
    In the FL setup, the goal is to learn a global model $f_\theta: \mathcal{X} \to \mathcal{Y}$, parametrized by $\theta$, on data distributed among $K$ clients without sharing local information. Each device $k\in[K]$ has access to $n_k$ samples from a local dataset $\mathcal{D}_k = \{x_i,y_i\}_{i=1}^{n_k}$ where $x\in\mathcal{X}$ is the input and $y\in\mathcal{Y}$ its corresponding label.
    
    FedAvg~\cite{mcmahan2017communication} follows an iterative approach based on $T$ communication rounds with the goal of solving
    \begin{equation}
        \arg\min_{\theta \in \mathbb{R}^d}\sum_{k\in \tilde{C}} \frac{n_k}{n} L_k(\theta), ~ d\in\mathbb{N^+}
        \label{eq:fedavg}
    \end{equation}
    where $L_k(\theta) = \mathbb{E}_{(x,y)\sim{\mathcal{D}_k}}[\ell_k(f_\theta;(x,y))]$ is the local empirical risk, $\ell_k$ the cross-entropy loss, and $n=\sum_{k} n_k$ the total amount of training data. At each round $t\in [T]$, the server sends $\theta_{t}$ to a fraction of $\tilde{C}$ randomly selected clients.
    Each client $k\in\tilde{C}$ computes its update $\theta_{t+1}^k$ using $\mathcal{D}_k$ by minimizing the local objective and sends it back to the server. The updated weights are then aggregated by the server into a new global model $f_{\theta_{t+1}}$ as:
    \begin{equation}
        \theta_{t+1} \leftarrow \sum_{k\in \tilde{C}} \frac{n_k}{n} \theta_{t+1}^k   
    \end{equation}
    
    However, in realistic scenarios, there is no guarantee that local datasets from different clients are drawn independently from the same underlying distribution.
    , \ie given two clients $i$ and $j$, $\mathcal{P}(\mathcal{D}_{i}) \neq \mathcal{P}(\mathcal{D}_{j})$. More in general, $f_{\theta^k} \neq f_\theta~\forall k$ clients~\cite{briggs2020federated}.
    In this work, we mitigate the issue of statistical heterogeneity in classification tasks by introducing \methodShort, an algorithm for FL that leverages sequential training among a sub-sample of clients $\tilde{C_S}$, grouped together according to their data distribution. Specifically, clients observing different data are grouped into a \textit{superclient} $S$ obtaining an approximation of the underlying uniform distribution over all $N_c$ classes, \ie $\bigcup_{k\in\tilde{C_S}} \mathcal{D}_k \sim \mathcal{U}_{[N_c]}$. Intuitively, thanks to the sequential training inside superclients, local models can accumulate knowledge on the majority of the classes even if single clients heavily heterogeneous. 
    
    \subsection{Building superclients}
    \label{subsec:superclients}
    Our goal is to build a superclient $S$ from users having different local distributions without breaking the privacy constraints, \ie without directly accessing the clients' data (Figure \ref{fig:fedseq_teaser}). We propose different grouping criteria $G_S$ as an ensemble of i) a \textit{client distribution approximator} $\psi_{(.)}$ providing statistics regarding the local distribution in a privacy-preserving way, ii) a \textit{metric} $\tau$ for evaluating the distance between the estimated data distributions and iii) a \textit{grouping method} $\phi_{(.)}$ to assemble dissimilar clients, \ie $G_S := \{\psi_{(.)}; \tau; \phi_{(.)}\}$.
    
    \subsubsection{\textbf{Client distribution approximator}}
    \label{subsec:approximator}
    We split the model $f_\theta$ into a deep feature extractor $h_{\theta_{\textnormal{feat}}}: \mathcal{X} \to \mathcal{Z}$ and a classifier $g_{\theta_{\textnormal{clf}}}: \mathcal{Z} \to \mathcal{Y}$, where $\theta = (\theta_{\textnormal{feat}}, \theta_{\textnormal{clf}})$ is the entire set of model parameters.
    The classification output is given by $g \circ h: \mathcal{X} \to \mathcal{Y}$, where we drop the subscripts to ease the notation.
    
    \methodShort~exploits a \textit{pre-training stage} to estimate the clients' data distribution, during which each client $k$ produces a model $f_{\theta_0^k}$ by training on its local dataset for $e$ epochs starting from the same random initialization $\theta_0$. We propose two strategies based on i) the parameters of the local classifier $\theta_{0,\textnormal{clf}}^k$ or ii) its predictions on a server-side public dataset $\mathcal{D}_{pub}$ $\{f^k(z) = g^k(h^k(z)),~z\in \mathcal{D}_{pub}\}$, respectively $\psi_{\textnormal{clf}}$ and $\psi_{\textnormal{conf}}$. As shown by~\cite{achille2019task2vec}, the model classifier is representative of the task it was trained on. That is why the weights of the classifier are used as proxy of the client's local distribution for $\psi_{\textnormal{clf}}$ and directly fed to the grouping method $\phi_{(.)}$.
    For $\psi_{\textnormal{conf}}$, we test each $f_{\theta_0^k}$ on a public \textit{\quotes{exemplar set}} $\mathcal{D}_{pub} = \bigcup_{c=1}^{N_c} \mathcal{D}_c$, where $\mathcal{D}_c$ contains $J$ samples for class $c\in[N_C]$. Then, we average the predictions by class as $p_{k,c}  = \frac{1}{J}\sum_{x \in \mathcal{D}_c} f_{\theta_0^k}(x)$, and 
    define the $k$-th client's \textit{confidence vector} as:
\begin{equation}
    p_k := \text{softmax}(\{p_{k,1},...,p_{k,N_C}\}) \in [0,1]^{N_C}
    \label{eq:conf_vectors}
\end{equation}
    Since the $k$-th model's predictions are favorable towards the majority of the classes seen in $\mathcal{D}_k$ \cite{he2009learning}, $p_k$ is an acceptable privacy-preserving representation of $\mathcal{D}_k$.
    In the following sections, we indicate as $\tilde{\mathcal{D}_k}$ the estimate provided by $\psi_{(.)}$ for the $k$-th device's data distribution.
    
    \subsubsection{\textbf{Grouping metrics}} 
    Starting from client $k$'s data approximation $\tilde{\mathcal{D}}_k$, we build similar superclients from users having different distributions, \ie we aim at minimizing the inter-superclients distance while maximizing the intra-superclient one. 
    To do so, given $\tilde{\mathcal{D}_{i}}$ and $\tilde{\mathcal{D}_{j}}$, we need a metric $\tau(\tilde{\mathcal{D}_{i}}, \tilde{\mathcal{D}_{j}}): \mathbb{R}^{N_C\times N_C} \to \mathbb{R}$ to measure the distance between the two distribution estimates. We choose the Euclidean distance and cosine similarity to compare the weights of the clients' classifier, 
    although other popular metrics can be used~\cite{villani2009optimal}.
    Alternatively, more sophisticated techniques learning features representation and cluster assignment can be used, such as \cite{Tian2017DeepClusterAG}. Since satisfying results were reached with the standard metrics (Table \ref{tab:ablation_clusters}), we have decided to favor a simpler and sufficiently effective approach.
    When $\tilde{\mathcal{D}_k}$ as the form of an actual probability distribution given by the confidence vector, we also adopt two \textit{disomogeneity} measures, the Gini index \cite{farris2010gini} and the Kullback-Leibler (KL) divergence \cite{kullback1951information}.
    \subsubsection{\textbf{Grouping method}} We first define $\mathcal{D}_S = \bigcup_{k\in\tilde{C_S}} \mathcal{D}_k$ as the union of the data from the clients $\tilde{C_S}$ belonging to a superclient $S$.
    Our aim is to find the maximum amount of superclients $N_S$ satisfying the following constraints: i) minimum number of samples $|\mathcal{D}_S|_{min}$ and ii) maximum number of clients $K_{S,max}$. 
    We introduce three strategies to find an approximation of the maximization problem,
    given the chosen $\psi_{(.)}$ and $\tau$.
    The first, $\phi_{\textnormal{rand}}$, is a naïve yet practical approach where clients are randomly assigned to superclients until the defined stopping criterion is met. The second one, $\phi_{\textnormal{kmeans}}$, is based on the K-means algorithm~\cite{steinley2006k}: first, K-means is applied to obtain $N_S$ homogeneous clusters; then, each superclient is formed by iteratively extracting one client at a time from each cluster, until the number of samples $|\mathcal{D}_S|$ in each superclient $S$ is at least $|\mathcal{D}_S|_{min}$ and the number of clients $K_{S} \leq K_{S,max}$ (detailed algorithm in Appendix \ref{app:grouping}). Lastly, $\phi_\textnormal{greedy}$ follows a greedy methodology to produce superclients. Initially, one random client $k_i$ is assigned to the current superclient $S$, $i \in [K]$. Then, the second client $k_j$ is chosen so as the distance between $k_i$ and $k_j$ is maximized, \ie $\max_{j \in [K]} \tau(\tilde{\mathcal{D}_{k_i}}, \tilde{\mathcal{D}_{k_j}})$. The process is repeated until the established maximum number of clients $K_{S,max}$ and the minimum number of samples $|\mathcal{D}_S|_{min}$ are reached by iteratively maximizing $\tau(\tilde{\mathcal{D}_{j}}, \frac{1}{|S|}\sum_{i\in |S|} \tilde{\mathcal{D}_{i}})$, with $|S|$ being the cardinality of $S$ until that point (see Appendix \ref{app:grouping}).

    \begin{algorithm}[t]
    \algsetup{linenosize=\tiny}
    \footnotesize
    \caption{\hlc[green!30!white]{\textsc{FedSeq}} and \hlc[blue!20!white]{\textsc{FedSeqInter}}}
    \label{alg:fedseq_fedseqInter}
    \begin{algorithmic}[1]
    \REQUIRE $f_{\theta_0}$, $G_S$, $K_{S,max}$, $|\mathcal{D}_S|_{min}$. Epochs $e$, $E_k$, $E_S$. $T$ rounds. Clients $K$. Fraction $C$ of superclients selected at each round.
    
    \STATE $S \leftarrow$ \textsc{CreateSuperclients}($f_{\theta_0}$, $G_S$, $e$, $K_{S,max}$, $|\mathcal{D}_S|_{min}$, $K$) 
    \STATE $N_S \leftarrow |S|$ 
    \STATE \hlc[blue!20!white]{$\Theta \leftarrow [\theta_0, \dots, \theta_0]_{1 \times CN_S}$, $w \leftarrow [0, \dots, 0]_{1\times CN_S}$} 
    \FOR{$t=0$ to $T$} 
        \STATE $S^t \leftarrow $ Subsample fraction $C$ of $N_S$ superclients
        \FOR{$S_i \in S^t$ \textbf{in parallel}}
            \STATE Shuffle clients in $S_i$
            \STATE \hlc[green!30!white]{$\theta^{S_i,0}_t \leftarrow \theta_t$} \hlc[blue!20!white]{$\theta^{S_i,0}_t \leftarrow \Theta[i]$}
            \FOR{$e_S=1$ to $E_S$}
                \STATE $\theta^{S_i}_{t+1} \leftarrow$ \textsc{SequentialTraining($\theta^{S_i,0}_t$, $E_k$)}
            \ENDFOR
        \STATE \hlc[blue!20!white]{$\Theta[i] \leftarrow \theta^{S_i}_{t+1}$,  $w_i \leftarrow w_i+ |\mathcal{D}_{S_i}|$}
        \ENDFOR
        \STATE  \hlc[green!30!white]{$\theta_{t+1} \leftarrow$ \textsc{FedAvg($\{\theta^{S_i}_{t+1},~\forall S_i \in S^t\}$)}}
        \IF{$t \mod N_S = 0$}
            \STATE \hlc[blue!20!white]{$\theta_{t+1} \leftarrow$ $\sum_i\frac{w_i}{w}\Theta[i]$, $w = \sum_i w_i$}
            \STATE \hlc[blue!20!white]{$\Theta \leftarrow [\theta_{t+1}, \dots \theta_{t+1}]$, $w \leftarrow [0, \dots, 0]$}
        \ENDIF
    \ENDFOR
    \end{algorithmic}
    \end{algorithm}

    \subsection{Sequential training}
    \subsubsection{\textbf{FedSeq}}
    
    Within each superclient $S_i$, with $i\in[N_S]$, training is performed in a sequential way, meaning that $S_i$ is considered as a sequence of clients $k_{i,1},\dots,k_{i,|S_i|}$. The server sends the global model $f_{\theta_t}$ to the first device $k_{i,1}$, which trains it for $E_k$ epochs 
    on $\mathcal{D}_{k_{i,1}}$. The obtained parameters $\theta^{k_{i,1}}_{t+1}$ are sent to the next client $k_{i,2}$. Such training procedure continues until the last client $k_{|S_i|}$ updates the received model, possibly repeating for $E_S$ times following a ring communication strategy.
    %
    Then, the last client sends its update to the server, where all the superclients updates are averaged according to Eq.~\ref{eq:fedavg}. The details of \methodShort~are summarized in Algorithm~\ref{alg:fedseq_fedseqInter}.
    
    \subsubsection{\textbf{FedSeqInter}}
    Sequentiality can be also exploited at a superclient level. At each round $t$, every selected superclient $S_i$ receives the model  $\theta^{S_j}_t$ from another previously involved superclient $S_j$, initially $\theta_0$. Every $N_S$ rounds the models are averaged, weighted by the number of examples on which each model was trained on. The insight behind this approach is that it might be useful to merge models only after they have been trained on a larger portion of the dataset. Statistically, after $N_S$ rounds, each model is likely to have been trained on the entire dataset, thus getting closer to a centralized scenario. This strategy requires far fewer aggregation and synchronization steps with the server: the possibility to go out of sync accounts for variance in clients' delays, allowing faster superclients not to be slacken by slower ones.

%% file: sections/4.experiments.tex
\section{Experiments}
\label{sec:experiments}
We evaluate \methodShort~on image classification tasks from  \textsc{Cifar-10} and \textsc{\textsc{Cifar-100}}, widely used as benchmarks in FL. In order to set up a heterogeneous scenario, the local class distribution is sampled from a Dirichlet distribution with $\alpha\in \{0, 0.2, 0.5\}$ \cite{hsu2019measuring}. Implementation details can be found in Appendix \ref{app:implementation}. We evaluate our results in terms of global accuracy on the test set (Tables \ref{tab:sota} and \ref{tab:fedseq_ablation}) and convergence rates (Table \ref{tab:rates}). All reported results are averaged over the last 100 rounds.

\subsection{Comparison with state-of-the-art FL algorithms}
\label{subsec:comparison}
We compare our method with the state-of-the-art (SOTA) algorithms FedAvg~\cite{mcmahan2017communication}, FedProx~\cite{li2020federated}, SCAFFOLD~\cite{karimireddy2020scaffold} and FedDyn~\cite{acar2021federated}. The analysis is presented both in terms of convergence performance (Figure~\ref{fig:exps_plots}, Table~\ref{tab:sota}) and speed (Table~\ref{tab:rates}). Taking into account both the convergence performance and rates, the best configuration chosen for FedSeq is based on the greedy grouping algorithm with KL-divergence applied on confidence vectors, \ie $G_S = \{\psi_\textnormal{conf}, \phi_\textnormal{greedy}, \tau_{KL}\}$. 
In addition, all results are compared with FedSeqInter, which adds the inter-superclient sequential training to this configuration, and is shown to outperform any configuration of FedSeq. 

\subsubsection{\textbf{Results at convergence}}
\input{tables/sotas}
\begin{figure*}[t]
\begin{center}
 \includegraphics[width=.9\linewidth]{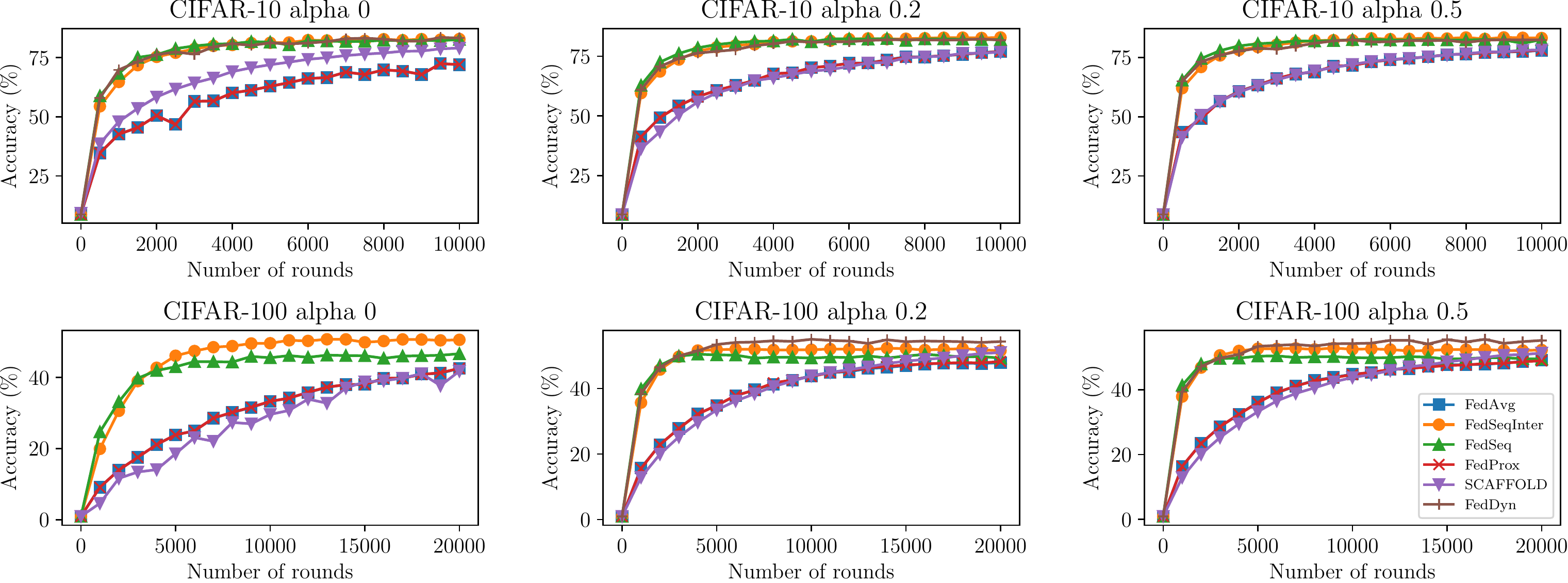}
 \end{center}
   \caption{Comparison between the results of SOTAs and the best configurations of FedSeq and FedSeqInter by varying $\alpha$ and dataset. FedSeqInter performs on par with FedDyn and both outperform the other approaches. Best viewed in color.}
 \label{fig:exps_plots}
 \end{figure*}

Table \ref{tab:sota} shows how \methodShort~reaches consistently better results than other methods not only when addressing extreme data heterogeneity, but also when faced with less severe conditions. This behavior reflects equally on both datasets. In particular, FedProx seems unable to address extreme scenarios, maintaining performances comparable to FedAvg. SCAFFOLD proves itself effective in addressing the most unbalance case ($\alpha = 0$), with $+7\%$ at convergence on \textsc{Cifar-10}, but fails at improving the results achieved by FedAvg both in more moderate scenarios and on \textsc{Cifar-100}. We found FedDyn to be the best current state-of-the-art algorithm, reaching the target accuracies for all configurations except \textsc{Cifar-100} with $\alpha=0$. \methodShort~successfully address the challenge of extremely unbalanced clients on both datasets, outperforming FedAvg, FedProx and SCAFFOLD both in terms of final performance and convergence speed, being on par with FedDyn in the average case (Figure \ref{fig:exps_plots}). FedSeqInter - although initially slower - reaches the highest accuracy value, close to that of the centralized scenario $Acc_{centr}$: in the most challenging setting, the achieved value corresponds to $96.4\%\cdot Acc_{centr}$ on \textsc{Cifar-10} and $91.2\%\cdot Acc_{centr}$ on \textsc{Cifar-100}.
That tells us that aggregating every $N_S$ rounds not only leads to less frequent synchronization between clients and server with a consequent speed up of the training process, but also improves the accuracy reached.
\subsubsection{\textbf{Integrating \methodShort\ with state-of-the-art}}
\label{subsec:integration_SOTA}
Since FedSeq keeps the same logic of FedAvg both in the local training and the server-side aggregation, it can be easily integrated with other approaches modifying those parts of the algorithm. In particular, we evaluate the performance of FedProx~\cite{li2020federated} and FedDyn~\cite{acar2021federated} on top of FedSeq, since changes to the local objective are straightforward to transfer in our sequential training framework.
FedProx adds a proximal term $\mu$ to the local objective to improve stability and regularize the distance between the local and global models.
We can repurpose FedProx to be used in our sequential framework by adding a proximal term to retain the information learned by the previous client rather than the global model, with potential benefits in the most challenging settings. 
Similarly FedDyn can be integrated in FedSeq by adding both linear and quadratic penalty terms to the loss function, using the model trained by the previous client in place of the server's last model (see Appendix \ref{app:SOTAintegration} for the details).
Results in Table~\ref{tab:sota} show that integrating FedSeq with FedProx makes the algorithm converge slightly faster only in the most unbalanced scenario, while performances are on par in the remaining the cases.

\subsubsection{\textbf{Convergence speed analysis}}

\input{tables/conv_rates}
In Table~\ref{tab:rates}, we report the time (indicated as number of rounds) needed by our best configurations and SOTAs to reach respectively the $70\%$, $80\%$ and $90\%$ of the centralized accuracy,
also indicating the speedup relative to FedAvg. Considering the most challenging situations, on \textsc{Cifar-10}, FedSeq based on the KL divergence on confidence vectors is $7$ times faster than FedAvg and successfully reaches the $90\%$ of the centralized accuracy in less than a third of rounds budget; on \textsc{Cifar-100}, FedSeqInter is the only algorithm able to reach the $90\%$ of the centralized accuracy, in less than half of the available rounds. As highlighted by \cite{varno2022minimizing}, we confirm that FedDyn is prone to parameters explosion in extremely imbalanced settings: when run on \textsc{Cifar-100} with $\alpha=0$, the model fails to converge. 


\subsection{Ablation study}
\label{subsec:ablation}

\input{tables/ablation_results}
In this Section, we provide information on the ablation studies performed on \methodShort. Specifically, the details regarding the pre-training phase and the construction of superclients are shown, together with the analysis of the different configurations available for \methodShort~which led to the choice presented in Section \ref{subsec:comparison}. 
\subsubsection{\textbf{Clients pre-training}}
All the  grouping criteria introduced in Section \ref{subsec:superclients} rely on the clients' data approximation $\tilde{\mathcal{D}}_k$, produced by the \textit{approximator} $\psi$. Regardless of the choice of $\psi$, the first step required for building superclients is a pre-training phase, local to every client. The randomly initialized model $f_{\theta_0}$ is trained by each device for $e$ epochs and is then exploited for estimating the data distribution without breaking the privacy constraints. 
Intuitively, $e$ should be large enough for the model to fit the local training set and at the same time as small as possible so as not to cause a computational burden on the clients. Hence we expect models trained on similar distributions to be more alike than those that have seen different ones. 
We tested 
$e \in \{1, 5, 10, 20, 30, 40\}$. For each of those values, we obtain the similarity matrix $D^e:=\{D^e_{ij}=\frac{\theta^i_e\cdot \theta^j_e}{||\theta^i_e|| \,||\theta^j_e||}\}$, representing the cosine distance between $f_{\theta^i_e}$ and $f_{\theta^j_e}$, where $\theta_e^i$ and $\theta_e^j$ are respectively the parameters of client $i$ and $j$ models trained for $e$ local epochs, $\forall (i,j)\in(K\times K)$. Figure \ref{fig:ablation:heat} shows those matrices as heatmaps for \textsc{Cifar-100} (see Appendix \ref{app:pre_train} for \textsc{Cifar-10}). In Figure \ref{fig:ablation:preTrainTrend}, the trend of $||D^e||$ for each value of $e$ is reported: we can notice how 5 epochs are sufficient for the models to be significantly different and after 10 epochs of pre-training the change rate of the models is reduced. Therefore, looking for the trade-off between the informative value of the trained models and the performance overhead, we choose $e=10$ as default value for the clients pre-training.

\subsubsection{\textbf{Estimating clients' data distribution}}
We can extract an estimate of the distribution of local datasets from the clients' pre-trained models via an approximator $\psi$ (see Sec.~\ref{subsec:approximator}. We compare $\psi_\textnormal{clf}$ and $\psi_\textnormal{conf}$, based respectively on the pre-trained classifier weights and on the confidence vectors (Eq. \ref{eq:conf_vectors}). As for the \textit{classifier} approximator, we test three scenarios: we use all three fully connected layers of the network, the last two or only the last.
To mitigate the \textit{curse of dimensionality}~\cite{bellman1966dynamic}, we apply PCA~\cite{pca} on the parameters, keeping 90\% of the explained variance. Our key
findings are that the percentage of preserved components: i) decreases with the complexity of the dataset, \ie less components are needed for \textsc{Cifar-10}, and ii) increases  directly proportional to $e$, except for $\alpha = 0$ (more details in Appendix \ref{app:ablation_clusters}).
We deduce that $10$ local epochs are already sufficient to capture the polarization of the dataset in its extreme imbalance. As for $\psi_\textnormal{conf}$, we retain $10$ images per class from the test set on the server-side ($\mathcal{D}_{pub}$) for testing the pre-trained models and computing the \textit{confidence vectors} as described in Section \ref{subsec:approximator}. Once $\mathcal{D}_{pub}$ has served its purpose, it is not used again.


\begin{figure}[]
\begin{center}
\subfloat[][\label{fig:ablation:heat}]{\includegraphics[ width=0.24\textwidth]{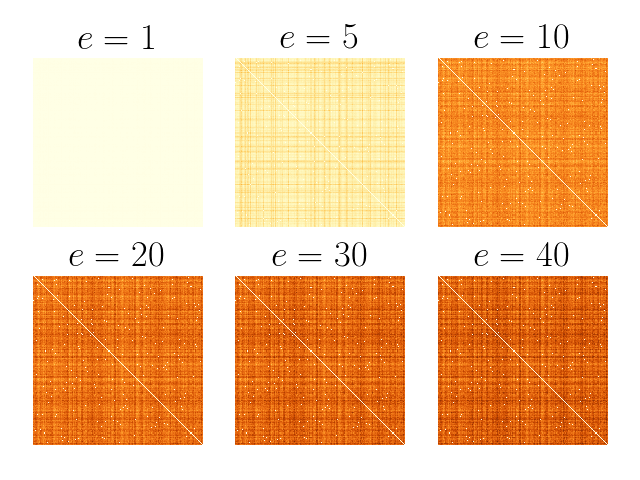}} 
\subfloat[][ \label{fig:ablation:preTrainTrend}]{\includegraphics[ width=0.24\textwidth]{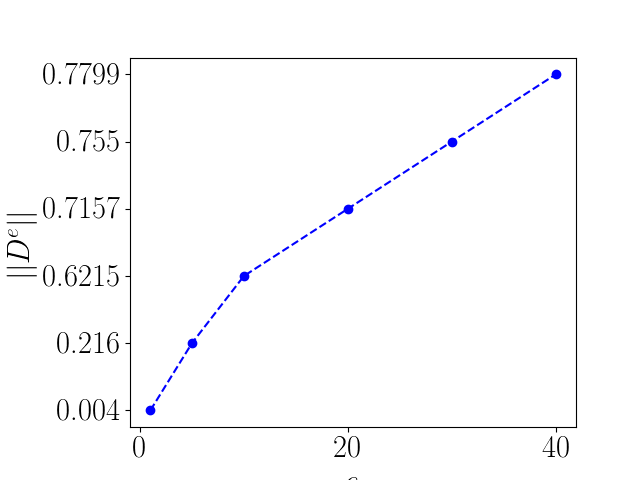}}
\end{center}
   \caption{Effect of pre-training $K=500$ local models for $e \in \{1, 5, 10, 20, 30, 40\}$ epochs on \textsc{Cifar-100}. (a) Heatmaps of the similarity matrix $D^e$. (b) Trend of $||D^e||$. After $e=10$ the slope of the curve decreases.}
\label{fig:preTrain}
\end{figure}


\subsubsection{\textbf{Comparison of grouping criteria}}
\label{subsubsec:grouping}
Here we provide the experimental results of the different combinations of grouping criteria $G_S$. A reasonable value of $K$ for $\phi_\textnormal{kmeans}$ 
is the number of classes of the dataset. To evaluate how \textit{homogenous} the superclients' overall data distribution is, we use the following measures:

\begin{itemize}
    \item \textit{balance ratio} $:= \frac{\min_{c\in[N_C]} N_c}{\max_{c\in[N_C]} N_c}$, where $N_{c}$ is the number of samples for the class $c$ 
    \item \textit{covered classes} $:= \frac{1}{N_C}\sum_{c=1}^{N_C} \mathds{1}_{P(y=c)>0}$.
\end{itemize} 
It is to be noted the percentage of classes covered is a less discriminatory measure, as the class is accounted for as present even if only one of its samples is in the superclient, while a low deviation from the mean of the samples per class is necessary to have a higher balance ratio, making the latter more reliable.
In Appendix \ref{app:ablation_clusters}, Table \ref{tab:ablation_clusters} shows the results varying by $\psi$, $\phi$ and $\tau$. The first consideration is the random assignment strategy ($\phi_\textnormal{rand}$) has surprisingly good indices.
The reason lies in statistical considerations on the setting: when $\alpha=0$, there are multiple clients (\ie 50 clients in \textsc{Cifar-10} and 5 in \textsc{Cifar-100}) having samples belonging to the same class; thus, a random choice in unlikely to group only those clients with the same data distribution. As $\alpha$ grows, each client has a more homogeneous distribution, so every clustering criterion leads to a similar result. $\phi_\textnormal{kmeans}$ has the best performances when $\alpha=0$, with zero variance on the number of clients in the same set, while $\phi_\textnormal{greedy}$ is the best one on average, 
becoming our algorithm of choice. Figures in Appendix \ref{app:superclients} show examples of superclients built with different $\phi$. 
As for the approximators, it is possible to see that, fixed the choice of $\phi_\textnormal{greedy}$, the use of $\psi_{\textnormal{clf}}$ leads to higher balance ratio, especially when $\tau_{\textnormal{cosine}}$ is adopted, while 
Table \ref{tab:fedseq_ablation} shows that 
$\psi_{\textnormal{conf}}$ brings towards higher accuracy. As for the metrics, the speedup with $\tau_{\textnormal{KL}}$ is more prominent (Table \ref{tab:rates}). So our approximator of choice is $\psi_{\textnormal{conf}}$ with $\tau_{\textnormal{KL}}$.

%% file: tables/sotas.tex
\setlength\dashlinedash{0.5pt}
\setlength\dashlinegap{1.5pt}
\setlength\arrayrulewidth{0.3pt}

\begin{table}[t]
    \begin{center}
    \caption{Comparison with SOTA FL algorithms.}
    \label{tab:sota}
    \resizebox{\linewidth}{!}{
    \begin{tabular}{c l c c c c}
        \toprule 
        Dataset & Algorithm & $\alpha  = 0$ & $\alpha  = 0.2$ & $\alpha  = 0.5$ & Centr.\\
        \midrule
        \multirow{9}{*}{\textsc{Cifar-10}} 
        & FedAvg & 71.41 & 76.82 & 77.98 & \multirow{9}{*}{85.72}\\
        & FedProx & 71.41 & 76.84 & 77.98&\\
        & SCAFFOLD & 79.02 & 76.47 & 78.25&\\
        & FedDyn &  \underline{\textbf{83.26}}	& 81.74 & 82.41& \\
        & \methodShort & 82.21 &	82.20 &	82.23 &\\
        & \methodShort Inter & 82.65 & \textbf{82.79} & \textbf{83.32}&\\
        \cdashline{2-5}
        & \methodShort~+ FedProx & 82.14 & 82.16 & 82.49&\\
        & \methodShort~+ FedDyn & 82.90 & \underline{\textbf{83.55}} &	\underline{\textbf{83.95}}&\\
        & \methodShort Inter + FedProx & 82.95 &	82.95 &	83.52&\\
        & \methodShort Inter + FedDyn &  \textbf{83.11} & 	83.06 &	83.33& \\
        \midrule
        \multirow{9}{*}{\textsc{Cifar-100}} 
        & FedAvg & 42.66 & 48.02 & 48.89&\multirow{9}{*}{55.13}\\
        & FedProx & 42.66 & 48.20 & 48.88&\\
        & SCAFFOLD & 42.04 & 51.04 & 51.20&\\
        & FedDyn & - &	\underline{\textbf{54.41}} & \underline{\textbf{54.99}} &\\
        & \methodShort & 46.00 &	49.55 &	49.82 &\\
        & \methodShort Inter & \textbf{50.27} & 51.60 & 51.94&\\
        \cdashline{2-5}
        & \methodShort~+ FedProx & 46.02 & 49.71 & 49.62&\\
        & \methodShort~+ FedDyn &  50.45 &	50.23 &	50.80 & \\
        & \methodShort Inter + FedProx & \textbf{\underline{51.13}} &	\textbf{51.54} &	52.33& \\
        & \methodShort Inter + FedDyn &  51.06	& 51.04	& \textbf{52.68}	& \\
        \bottomrule
    \end{tabular}}
    \end{center}
\end{table}

%% file: tables/conv_rates.tex
\setlength\dashlinedash{0.5pt}
\setlength\dashlinegap{1.5pt}
\setlength\arrayrulewidth{0.3pt}

\begin{table*}[]
\centering
\caption{Convergence rates for the best configurations of FedSeq (1: $\{ \psi_\textnormal{conf}, \phi_\textnormal{greedy}, \tau_{KL}\}$, 2: $\{ \psi_\textnormal{clfAll}, \phi_\textnormal{greedy}, \tau_{cosine}\}$) and SOTAs. We report the round in which the 70\%, 80\% and 90\% of centralized accuracy is reached (\quotes{--} if the target accuracy was not reached), together with the speedup relative to FedAvg (\quotes{--} if FedAvg did not reach the target accuracy).}
\label{tab:rates}
\resizebox{\linewidth}{!}{
    \begin{tabular}{lllllllllll}
    \toprule
    \multicolumn{1}{c}{\multirow{2}{*}{Dataset}} &
      \multirow{2}{*}{Method} &
      \multicolumn{3}{c}{$\alpha  = 0$} &
      \multicolumn{3}{c}{$\alpha  = 0.2$} &
      \multicolumn{3}{c}{$\alpha  = 0.5$} \\
    \cmidrule(lr){3-5} \cmidrule(lr){6-8} \cmidrule(lr){9-11} &
       &
      \multicolumn{1}{c}{70\%} &
      \multicolumn{1}{c}{80\%} &
      \multicolumn{1}{c}{90\%} &
      \multicolumn{1}{c}{70\%} &
      \multicolumn{1}{c}{80\%} &
      \multicolumn{1}{c}{90\%} &
      \multicolumn{1}{c}{70\%} &
      \multicolumn{1}{c}{80\%} &
      \multicolumn{1}{c}{90\%} \\
     \midrule
     
    \multirow{10}{*}{CIFAR-10} &
    FedAvg                           & 4036 (1x)    & 7649 (1x)    & - (--)    & 2384 (1x)    & 4507 (1x)    & - (--)    & 1945 (1x)    & 3749 (1x)   & 8791 (1x)    \\
    & FedProx                          & 4036 (1x)    & 7649 (1x)    & - (--)    & 2384 (1x)    & 4507 (1x)    & - (--)    & 1946 (1x)    & 3753 (1x)   & 8786 (1x)    \\
    & SCAFFOLD                         & 2229 (1,81x) & 3914 (1,95x) & 8043 (--) & 2554 (0,93x) & 4771 (0,94x) & - (--)    & 1934 (1,01x) & 3761 (1x)   & 8453 (1,04x) \\
    & FedDyn &
      {\textbf{563 (7,17x)}} &
      \textbf{954 (8,02x)} &
      \textbf{2131 (--)} &
      \textbf{450 (5,3x)} &
      \textbf{797 (5,65x)} &
      \textbf{2059 (--)} &
      \textbf{374 (5,2x)} &
      \textbf{634 (5,91x)} &
      \textbf{1648 (5,33x)} \\
  
    \cmidrule(lr){2-2} &
     FedSeq\textsuperscript{1} &
      594 (6,79x) &
      991 (7,72x) &
      2047 (--) &
      407 (5,86x) &
      746 (6,04x) &
      1682 (--) &
      325 (5,98x) &
      619 (6,06x) &
      1358 (6,47x) \\
    & FedSeq\textsuperscript{2} &
      873 (4,62x) &
      1502 (5,09x) &
      3677 (--) &
      387 (6,16x) &
      720 (6,26x) &
      1543 (--) &
      323 (6,02x) &
      620 (6,05x) &
      1409 (6,24x) \\
    & FedSeq\textsuperscript{1} + FedProx & 594 (6,79x)  & 991 (7,72x)  & 2046 (--) & 407 (5,86x)  & 746 (6,04x)  & 1682 (--) & 325 (5,98x)  & 619 (6,06x) & 1358 (6,47x) \\
    & FedSeq\textsuperscript{1} + FedDyn  & {\ul {\textbf{345 (11,7x)}}} &
  {\ul {\textbf{581 (13,17x)}}} &
  {\ul {\textbf{1341 (--)}}} &
  {\ul {\textbf{253 (9,42x)}}} &
  {\ul {\textbf{447 (10,08x)}}} &
  {\ul {\textbf{1113 (--)}}} &
  {\ul {\textbf{232 (8,38x)}}} &
  {\ul {\textbf{403 (9,3x)}}} &
  {\ul {\textbf{933 (9,42x)}}}\\
     \cmidrule(lr){2-2} &
     FedSeqInter\textsuperscript{1}           & 762 (5,3x)   & 1305 (5,86x) & 2492 (--) & \textbf{538 (4,43x)}  & 1004 (4,49x) & 2099 (--) & \textbf{433 (4,49x)}  & \textbf{814 (4,61x)} & 1805 (4,87x) \\
    & FedSeqInter\textsuperscript{1} + FedProx &   735 (5,49x) &	1264 (6,05x) &	2388 (--) &	544 (4,38x)	& 1000 (4,51x) &	2084 (--) &	436 (4,46x)	& 825 (4,54x) &	\textbf{1747 (5,03x)}\\
    & FedSeqInter\textsuperscript{1} + FedDyn &  \textbf{733 (5,51x)} &	\textbf{1262 (6,06x)}&	\textbf{2344 (--)}&	\textbf{533 (4,47x)}&	\textbf{959 (4,7x)}&	\textbf{2061 (--)}&	\textbf{425 (4,58x)}&	\textbf{796 (4,71x)}	& 1750 (5,02x)\\
    \midrule 
    \multirow{10}{*}{CIFAR-100} &
    FedAvg &
      14412 (1x) &
      - (--) &
      - (--) &
      6409 (1x) &
      10253 (1x) &
      - (--) &
      5879 (1x) &
      9331 (1x) &
      - (--) \\
    & FedProx &
      14412 (1x) &
      - (--) &
      - (--) &
      6363 (1,01x) &
      10277 (1x) &
      - (--) &
      5918 (0,99x) &
      9250 (1,01x) &
      - (--) \\
    & SCAFFOLD &
      14483 (1x) &
      - (--) &
      - (--) &
      7088 (0,9x) &
      10191 (1,01x) &
      17200 (--) &
      6951 (0,85x) &
      10373 (0,9x) &
      16744 (--) \\
    & FedDyn &
      {\ul \textbf{- (--)}} &
      {\ul \textbf{- (--)}} &
      \textbf{- (--)} &
      \textbf{1031 (6,22x)} &
      \textbf{1603 (6,4x)} &
      \textbf{2634 (--)} &
      \textbf{868 (6,77x)} &
      \textbf{1433 (6,51x)} &
      \textbf{3018 (--)} \\
  
    \cmidrule(lr){2-2} & 
    FedSeq\textsuperscript{1} &
      3009 (4,79x) &
      5780 (--) &
      - (--) &
      {\ul \textbf{901 (7,11x)}} &
      1421 (7,22x) &
      \textbf{3436 (--)} &
      854 (6,88x) &
      {\ul \textbf{1264 (7,38x)}} &
      {\ul \textbf{2812 (--)}} \\
    & FedSeq\textsuperscript{2} &
      3968 (3,63x) &
      9378 (--) &
      - (--) &
      {\ul \textbf{922 (6,95x)}} &
      {\ul \textbf{1396 (7,34x)}} &
      {\ul \textbf{3924 (--)}} &
      {\ul \textbf{843 (6,97x)}} &
      1266 (7,37x) &
      {\ul \textbf{2713 (--)}} \\
    & FedSeq\textsuperscript{1} + FedProx &
      2946 (4,89x) &
      6005 (--) &
      - (--) &
      898 (7,14x) &
      1397 (7,34x) &
      3033 (--) &
      843 (6,97x) &
      1298 (7,19x) &
      2833 (--) \\
    & FedSeq\textsuperscript{1} + FedDyn &
       {\ul {\textbf{1914 (7,53x)}}} &
          {\ul {\textbf{3293 (--)}}} &
          {\ul {\textbf{7511 (--)}}} &
          {\ul {\textbf{556 (11,53x)}}} &
          {\ul {\textbf{987 (10,39x)}}} &
          {\ul {\textbf{2014 (--)}}} &
          {\ul {\textbf{541 (10,87x)}}} &
          {\ul {\textbf{957 (9,75x)}}} &
          {\ul {\textbf{1912 (--)}}} \\
     \cmidrule(lr){2-2} &
     FedSeqInter\textsuperscript{1} &
      3028 (4,76x) &
      4333 (--) &
      8494 (--) &
      1177 (5,45x) &
      1734 (5,91x) &
      3004 (--) &
      1034 (5,69x) &
      \textbf{1524 (6,12x)} &
      2675 (--) \\
    & FedSeqInter\textsuperscript{1} + FedProx &
       3027 (4,76x) &	4310 (--) &	\textbf{7149 (--)} &	\textbf{1163 (5,51x)} &	\textbf{1721 (5,96x)} &	\textbf{2915 (--)} &	1033 (5,69x) &	1525 (6,12x) &	2616 (--) \\
    & FedSeqInter\textsuperscript{1} + FedDyn &  \textbf{2964 (4,86x)}&	\textbf{4183 (--)}&	7539 (--)&	1180 (5,43x)&	1757 (5,84x)&	3018 (--)&	\textbf{1031 (5,7x)}&	1538 (6,07x)&	\textbf{2547 (--)}\\
     \bottomrule \\
    \end{tabular}
}
\end{table*}

%% file: tables/ablation_results.tex
\setlength\dashlinedash{0.5pt}
\setlength\dashlinegap{1.5pt}
\setlength\arrayrulewidth{0.3pt}

\begin{table}[t]
    \begin{center}
    \caption{FedSeq baselines: comparison of grouping criteria by varying $\phi$, $\psi$ and $\tau$. Results in terms of accuracy (\%).}
    \label{tab:fedseq_ablation}
    \resizebox{\linewidth}{!}{
    \begin{tabular}{l l l l l c c }
        \toprule
        Method & $\psi$ & $\phi$ & $\tau$ & $\alpha  = 0$ & $\alpha  = 0.2$ & $\alpha  = 0.5$ \\
        \midrule
        \multicolumn{7}{c}{\textbf{\textsc{Cifar-10}}} \\
        \midrule
        \multirow{7}{*}{\methodShort}
        & - & random & - & 81.90 & 82.09 & 82.12\\
        & clf & K-means & Euclidean & \textbf{82.30} & 81.78 & 82.48\\
        & conf & K-means & Euclidean & 82.04 & 81.99 & 82.37\\
        & conf & greedy & KL & 82.21 & \textbf{82.20} & 82.22 \\
        & conf & greedy & Cosine & 82.09 & 81.85 & 82.71 \\
        & clf & greedy & Cosine & 79.95 & 82.06 & \textbf{82.83}\\
        \cdashline{2-7}
        \methodShort Inter & conf & greedy & KL & \underline{\textbf{82.65}} & \underline{\textbf{82.79}} & \underline{\textbf{83.32}}\\
        \midrule
        \multicolumn{7}{c}{\textbf{\textsc{Cifar-100}}} \\
        \midrule
        \multirow{7}{*}{\methodShort} & - & random & - & \textbf{46.39} & 48.62 & 49.44\\
        & clf & K-means & Euclidean & 44.91 & 48.74 & 49.60\\
        & conf & K-means & Euclidean & 43.55 & 49.43 & 49.79\\
        & conf & greedy & KL & 45.97 & \textbf{49.56} & \textbf{49.82} \\
        & conf & greedy & Cosine & 45.79 & 48.98 & 49.61 \\
        & clf & greedy & Cosine & 45.22 & 48.92 & 49.62\\
        \cdashline{2-7}
        \methodShort Inter & conf & greedy & KL & \underline{\textbf{50.27}} & \underline{\textbf{51.60}} & \underline{\textbf{51.94}}\\
        \bottomrule
    \end{tabular}}
    \end{center}
\end{table}

%% file: sections/5.conclusion.tex
\section{Conclusion} 
In this work we address statistical heterogeneity in FL introducing \methodShort, the first approach exploiting sequential training of clients grouped by data dissimilarity (\textit{superclients}). We evaluate different stategies for grouping clients, based on privacy-preserving approximations of their local distributions, and 
show that FedSeq is robust to suboptimal solutions. 
We extend sequential training to superclients to reduce the impact of slow devices (FedSeqInter) and find the convergence performances improve.
Our comparative analysis with the state-of-art shows FedSeq largely outperforms FedAvg, FedProx and SCAFFOLD in terms of convergence accuracy and speed on both extreme and less severe non-i.i.d. scenarios, 
while performing on par with FedDyn on average.
Finally, empirical  results show  that  combining  existing  algorithms  with  FedSeq further improves its final performance and convergence speed.

%% file: supplementary.tex



\subsection{\textsc{Grouping Algorithms}}
\label{app:grouping}
Here, we provide details on the grouping algorithms described in Section \ref{sec:method}.

\begin{algorithm}
\caption{K-means grouping method $\phi_\textnormal{kmeans}$}
\label{alg:k-means cluster}
\begin{algorithmic}[1]
\REQUIRE $K$ clients, $\{\tilde{\mathcal{D}_1}, \dots, \tilde{\mathcal{D}_K}\}$ clients' approximated distributions, $|\mathcal{D}_S|_{min}$ minimum number of samples per superclient, $K_{S,max}$ maximum number of clients per superclient, grouping metric $\tau$, $n_k$ number of images on $k$th device

\STATE $N = \textnormal{number of classes}$
\STATE $C_1,\dots, C_N \: =\: \textsc{K-means}(\{\tilde{\mathcal{D}_1}, \dots, \tilde{\mathcal{D}_K}\},\tau, N)$ \COMMENT{K-means algorithm with $K=N$ returns $N$ homogeneous clusters}
\STATE $z \leftarrow 0, ~S = [\:], ~j=0$ \COMMENT{with $S$ being the set of superclients and $z$ its index}
\WHILE{$|\bigcup_{i=1}^N C_i|>0$}
    \STATE $S_z \leftarrow [\: ]$, $N_z \leftarrow 0$
    \WHILE{$N_z<|\mathcal{D}_S|_{min} ~\textbf{and}~ |S_z|<K_{S,max}$}
        \STATE $k \leftarrow \textsc{random}(C_j)$
        \STATE $S_z.\textsc{add}(k)$, $C_j.\textsc{remove}(k)$
        \STATE $j \leftarrow ((j+1)\mod N) $
        \STATE $N_z \leftarrow N_z+n_k$
    \ENDWHILE
    \STATE $S.\textsc{add}(S_z)$
    \STATE $z \leftarrow z+1$
\ENDWHILE
\RETURN $S$
\end{algorithmic}
\end{algorithm}

\begin{algorithm}
\caption{Greedy grouping method $\phi_\textnormal{greedy}$}
\label{alg:greedy cluster}
\begin{algorithmic}[1]
\REQUIRE $K$ clients, $\{\tilde{\mathcal{D}_1}, \dots, \tilde{\mathcal{D}_K}\}$  clients' approximated distributions, $|\mathcal{D}_S|_{min}$ minimum number of samples per superclient, $K_{S,max}$ maximum number of clients per superclient, grouping metric $\tau$, $n_k$ number of images on $k$th device

\STATE $z \leftarrow 0,\: S = [\:],\: \tilde{K}\leftarrow [k_1,\dots k_K]$
\WHILE{$|\tilde{K}|>0$}
    \STATE $S_z \leftarrow [\: ], \: N_z \gets 0$
    \STATE $k_i \leftarrow \textsc{random}(\tilde{K}) $
    \STATE $S_z.\textsc{add}(k_i)$, $\tilde{K}.\textsc{remove}(k_i)$
    \STATE $\tilde{\mathcal{D}}_{S_z} \leftarrow \tilde{\mathcal{D}}_i, \: N_z \gets N_z+n_i$
    \WHILE{$N_z<|\mathcal{D}_S|_{min} ~\textbf{and}~ |S_z|<K_{S,max}$}
        \STATE $k_j \leftarrow argmax_j(\tau(\tilde{\mathcal{D}}_j,\tilde{\mathcal{D}}_{S_z}))$
        \STATE $\tilde{\mathcal{D}}_{S_z} \leftarrow \frac{1}{2}\tilde{\mathcal{D}}_{S_z}+\frac{1}{2}\tilde{\mathcal{D}}_j$
        \STATE $N_z \leftarrow N_z+n_j$
        \STATE $S_z.\textsc{add}(k_j)$, $\tilde{K}.\textsc{remove}(k_j)$
    \ENDWHILE
    \STATE $S.\textsc{add}(S_z)$
    \STATE $z \leftarrow z+1$
\ENDWHILE
\RETURN $S$
\end{algorithmic}
\end{algorithm}

\subsection{\textsc{Clients' pre-training on \textsc{Cifar-10}}}
\label{app:pre_train}
\begin{figure}[]
\begin{center}
\subfloat[][\label{fig:ablation:heat}]{\includegraphics[ width=0.4\textwidth]{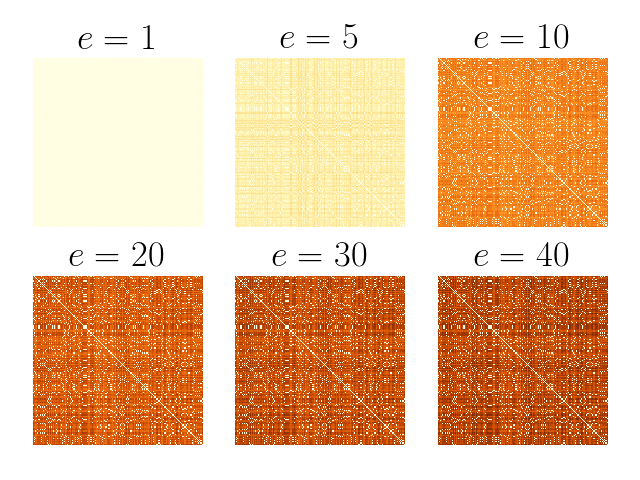}} \quad
\subfloat[][ \label{fig:ablation:preTrainTrend}]{\includegraphics[ width=0.4\textwidth]{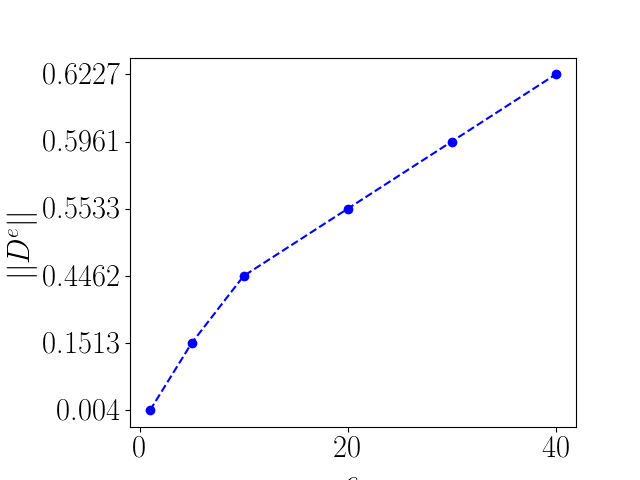}}\quad
\end{center}
   \caption{Effect of pre-training $K=500$ local models for $e \in \{1, 5, 10, 20, 30, 40\}$ epochs on \textsc{Cifar-10}. \textbf{(a)} Heatmaps of the similarity matrix $D^e$. \textbf{(b)} Trend of $||D^e||$. After $e=10$ the slope of the curve decreases.}
\label{fig:preTrain}
\end{figure}

Figure \ref{fig:preTrain} shows the effect of pre-training local models varying the number of local epochs $e$ for \textsc{Cifar-10}. As shown for \textsc{Cifar-100} in the main paper, we find a trade-off between the informative value of the trained models and the performance overhead with $e=10$.  
The results obtained are consistent across both datasets, showing that the chosen network is able to correctly fit both of them. 

\subsection{\textsc{Comparison of grouping criteria}}
\label{app:ablation_clusters}
\begin{figure}[h]
\begin{center}
\includegraphics[width=0.5\linewidth]{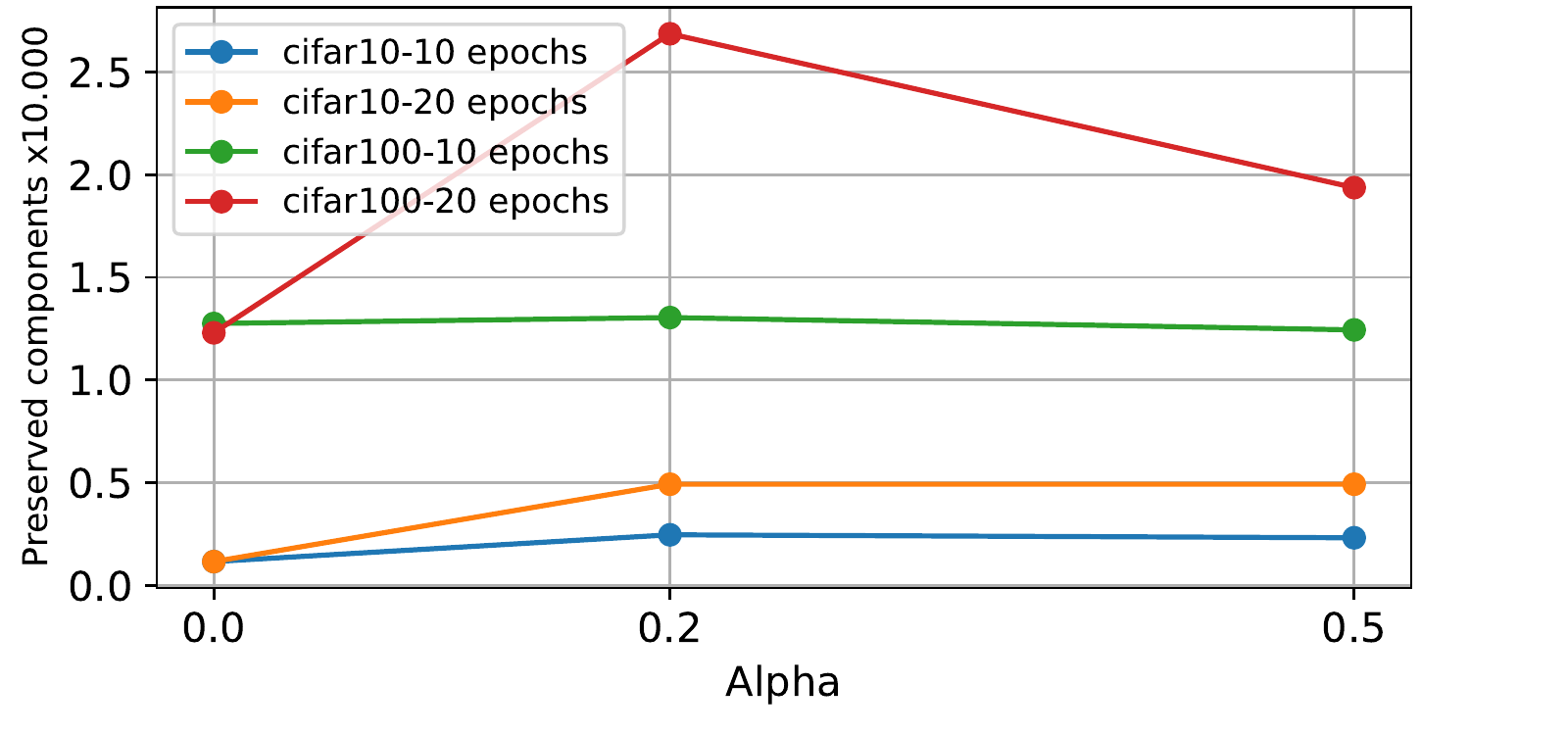}
\end{center}
   \caption{Ratio of the preserved components after applying PCA with 90\% of explained variance.}
\label{fig:ablation:PCA}
\end{figure}

Table \ref{tab:ablation_clusters} shows experimental results of the different combinations of grouping criteria $G_S$. We remind that the goal of our approach is to group clients with different distributions in the same superclient, in order to obtain heterogeneous ones. To this end, our evaluation metrics are the \textit{balance ratio} and \textit{covered classes} (see Section \ref{subsubsec:grouping} of the main paper), as a way to reflect the heterogeneity of superclient's dataset. 
The approximators \textit{classifierAll}, \textit{classifierLast2} and \textit{classifierLast} refer respectively to extracting the weights of all, the last two or only the last fully connected layer from our network of choice, LeNet-5. In practice, since we apply PCA on the network parameters (Figure \ref{fig:ablation:PCA}), extracting all the classifier's weights does not introduce much additional computational burden. Moreover, \textit{classifierAll} achieves the best performance among the three options. Therefore we choose to always extract all the weights.
Results are consistent across the dataset and show that the best combinations are $G^a_S = \{\psi_\textnormal{clf}, \phi_\textnormal{greedy}, \tau_{cosine}\}$ and $G^b_S = \{\psi_\textnormal{conf}, \phi_\textnormal{greedy}, \tau_{cosine}\}$. Experimental results on the performance of FedSeq given such grouping criteria show that on the average case $G^c_S = \{\psi_\textnormal{conf}, \phi_\textnormal{greedy}, \tau_{KL}\}$ leads to best results (see Section \ref{subsubsec:grouping}).
\input{tables/ablation_cluster}

\subsection{\textsc{Superclients analysis}}
\label{app:superclients}
\begin{figure}[h]
    \begin{center}
    \includegraphics[width=.9\linewidth]{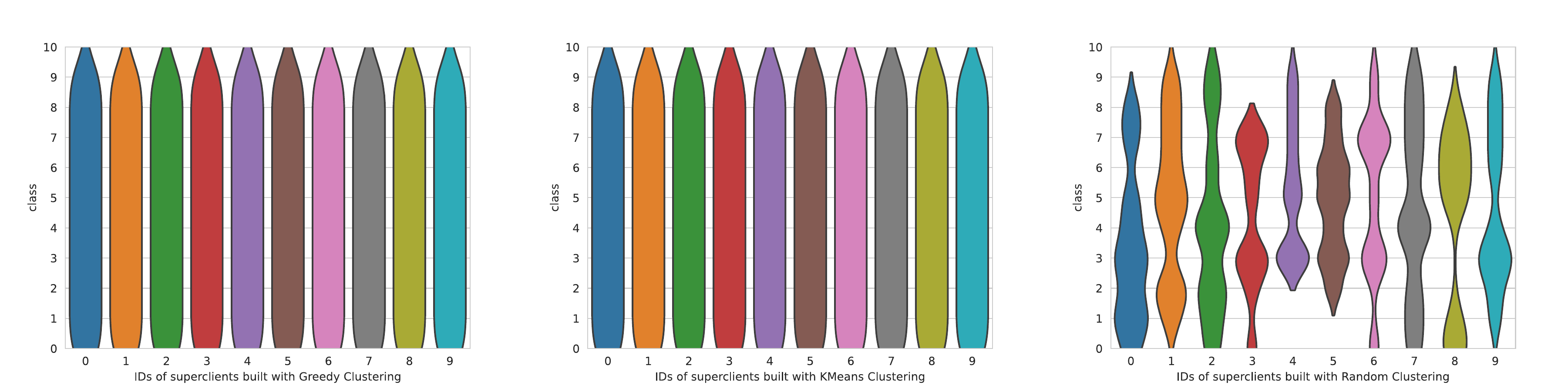}
    \end{center}
       \caption{Example of superclient distributions produced by different grouping algorithms on \textsc{Cifar-10} and $\alpha=0$.}
    \label{fig:violin-cluster-cifar10}
\end{figure}

\begin{figure}[h]
    \begin{center}
    \includegraphics[width=.9\linewidth]{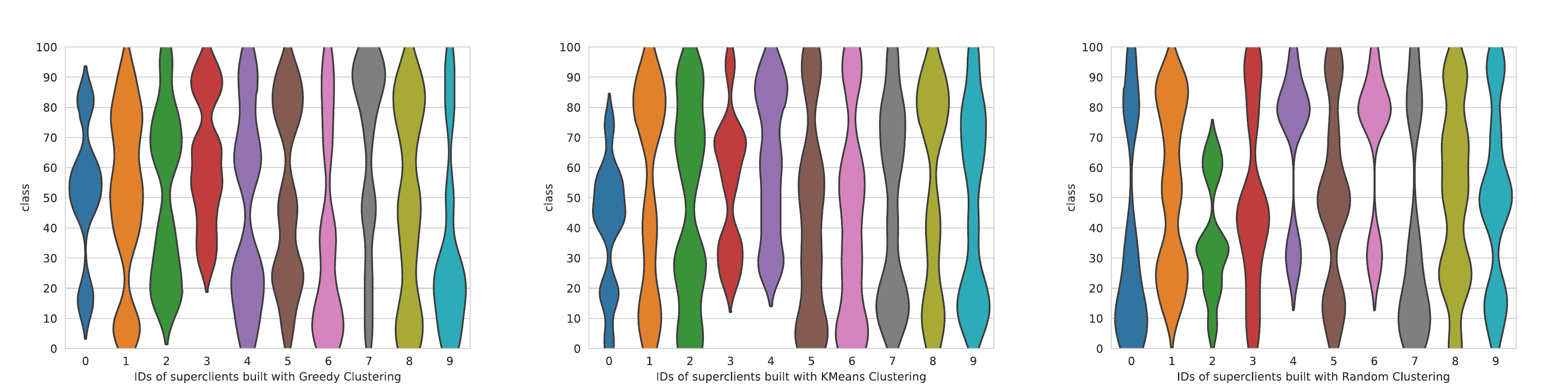}
    \end{center}
       \caption{Example of superclient distributions produced by different grouping algorithms on \textsc{Cifar-100} and $\alpha=0$.}
    \label{fig:violin-cluster-cifar100}
\end{figure}

\begin{figure}[]
    \begin{center}
    \includegraphics[width=.9\linewidth]{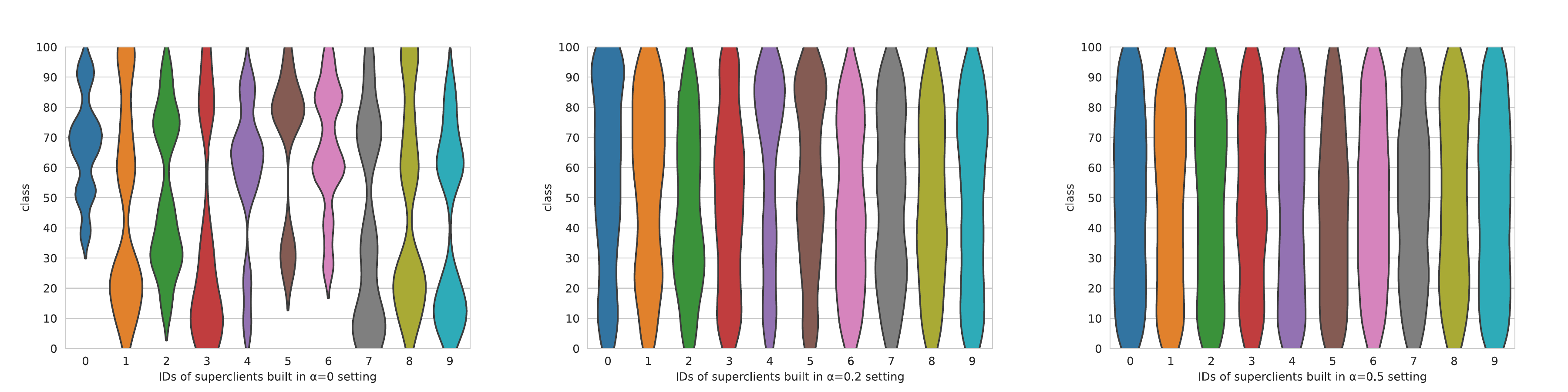}
    \end{center}
       \caption{Example of superclient distributions in different $\alpha$ settings with $\phi_{rand}$.}
    \label{fig:violin-cluster-alpha}
\end{figure}

Figures \ref{fig:violin-cluster-cifar10},\ref{fig:violin-cluster-cifar100},\ref{fig:violin-cluster-alpha} show superclients distributions in different settings. Figure \ref{fig:violin-cluster-cifar10} represents the distribution of 10 superclients built, from left to right, with $\phi_{greedy}$, $\phi_{kmeans}$ and $\phi_{rand}$ on \textsc{Cifar-10}. It is clear that the first two methods are able to build perfectly homogeneous superclients, while $\phi_{rand}$ struggles in doing so. Figure \ref{fig:violin-cluster-cifar100} shows the same configuration on \textsc{Cifar-100}: in this case, the advantage of using $\phi_{greedy}$ or $\phi_{kmeans}$ over $\phi_{rand}$ is not as evident, but the superclient distributions created with the first two clustering methods are still spread more homogeneously over the classes. Figure \ref{fig:violin-cluster-alpha} demonstrates the effect of $\alpha$ (from left to right: $0$, $0.2$ and $0.5$) in the construction of the superclients: the bigger the value of $\alpha$, the more homogeneous the superclients distributions are, regardless of the clustering method.

\subsection{\textsc{Implementation details}}
\label{app:implementation}
We evaluate \methodShort~on image classification tasks on two synthetic datasets widely used as benchmarks in FL, namely \textsc{Cifar-10} and \textsc{\textsc{Cifar-100}}. 
As for the data partitioning, we follow the protocol described in \cite{hsu2019measuring}: the class distribution of every client is sampled from a Dirichlet distribution with varying concentration parameter $\alpha$. 
Since our method addresses statistical heterogeneity, in our experiments we use $\alpha \in \{0, 0.2, 0.5\}$ that, combined with the number of clients $K$ among which the dataset is split ($K=500$), sets up a realistic scenario in which clients have small and very unbalanced datasets. 

Accounting for the difficulty of the task, we run the experiments for $T=10k$ rounds on \textsc{Cifar-10} and $20k$ on \textsc{Cifar-100}. The fraction of clients selected at each round is $C=0.2$. Following the setup of \cite{hsu2020federated}, our model is their proposed version of \textit{LeNet-5}, with a client learning rate of $0.01$, weight decay set to $4 \cdot 10^{-4}$, momentum $0$ and batch size $64$. 
As for the centralized scenario, we add a momentum of $0.9$ and a cosine annealing schedule for the learning rate, training the model for $300$ epochs. As for the clustering methods, we fix $|\mathcal{D}_S|_{min}=800$ and  $K_{S,max}=11$. In \methodShort, we fix $E_k=E_S = 1$ and similarly $E=1$ for FedAvg and the other SOTAs. An analysis on the choice of $E_S$ can be found in Appendix F: we show it is not convenient to perform more than one epoch through a superclient. 
For FedProx we evaluate $\mu\in\{10^{-4}, 10^{-3}, 10^{-2}\}$ and choose $\mu=0.01$, while for FedDyn $\alpha_{dyn}=0.1$ is chosen from the finetuning set $\{10^{-3}, 10^{-2}, 10^{-1}\}$.

Regarding FedDyn, we were unable to obtain the results for \textsc{Cifar-100} with $\alpha=0$: we conjecture that in our setting the amount of local update was not enough to calculate the appropriate $h^t$ server side, and the model diverged. To confirm this intuition we successfully ran the same case using a learning rate of $0.1$; the same happens when integrating FedDyn with FedSeq, in which case the models has more updated before returning to the server for the aggregation.

When integrating FedProx in FedSeq, we use $\mu=0.01$ chosen from $\{10^{-4}, 10^{-3}, 10^{-2}\}$; when instead we integrate in FedSeqInter, we choose $\mu=1$ chosen from $\{10^{-2}, 10^{-1}, 1, 10\}$: the rationale behind having selected higher values is that in a sequential training with loose aggregation it can be beneficial to try to retain more knowledge from the previous client's training. The experimental results in section \ref{sec:experiments} of the main paper confirm this intuition.
Similarly when integrating FedDyn in FedSeq, we use $\alpha_{dyn}=0.1$ chosen from $\{10^{-3}, 10^{-2}, 10^{-1}\}$, while when integrating in FedSeqInter we choose $\alpha_{dyn}=1$ from $\{10^{-2}, 10^{-1}, 1\}$.

\subsection{\textsc{Details on the integration of FedSeq with state-of-the-art}}
\label{app:SOTAintegration}

\vspace{2.5pt}
\textit{\textbf{FedProx}}
\vspace{2.5pt}

As pointed out in section \ref{subsec:integration_SOTA}, FedProx  adds  a  proximal  term $\mu$ to  the local objective to improve stability and regularize the distance between  the  local  and  global  models, modifying the local objective function as follows:

\begin{equation}
    \theta_k^t = \argmin_\theta(R_k(\theta; \theta^{t-1})= L_k(\theta) + \frac{\mu}{2} ||\theta -\theta^{t-1}||^2)
\end{equation}

In our setting, incorporating FedProx objecting function into the sequential training means trying to retain the information learned by the previous client rather than the global model, with potential benefits in the most challenging settings. In fact, because when $\alpha=0$ clients have local dataset with samples belonging only to one class, adding a proximal term could help avoiding the model shift towards the new learned task. In such a case, the objective function becomes:

 \begin{equation}
    \theta_{S_{k,j}}^t = \argmin_\theta(R_{S_{k,j}}(\theta; \theta_{S_{k,j-1}}^t)= L_{S_{k,j}}(\theta) + \frac{\mu}{2} ||\theta -\theta_{S_{k,j-1}}^t||^2)
\end{equation}
where $\theta_{S_{k,j-1}}^t$ is the model after the training of client $j-1$ belonging to superclient $S_k$.

\vspace{5.5pt}
\textit{\textbf{FedDyn}}
\vspace{2.5pt}

In FedDyn the proposed risk objective dynamically modifies local loss functions, so that, if in fact local models converge to a consensus, the consensus point is consistent with stationary point of the global loss\cite{acar2021federated}. Namely, each device computes:
  \begin{equation}
    \theta_k^t = \argmin_\theta(R_k(\theta; \theta_k^{t-1}, \theta^{t-1})= L_k(\theta) + \langle \nabla L_k(\theta_k^{t-1})\,,\theta\rangle +
    \frac{\alpha_{dyn}}{2} ||\theta -\theta^{t-1}||^2)
\end{equation}
FedDyn authors point out that for the first order condition for local optima to be satisfied, as $\theta_k^t \to \theta_k^\infty$ and $\nabla L_k (\theta_k^t) \to \nabla L_k (\theta_k^\infty)$, $\theta^t \to \theta_k^\infty$ which implies $\theta_k^\infty \to \theta^\infty$.
Then the server side aggregation updates the model such that:
\begin{equation}
    \theta^t= \frac{1}{|P_t|} \sum_{k \in P_t} \theta_k^t - \frac{1}{m} \sum_{k \in P_t}(\theta_k^t - \theta^{t-1})
\end{equation}
being $P_t$ the subset of client selected at round $t$. In this way $\theta^t$ convergence implies $h^t \to 0$.

When incorporating it in FedSeq, the dynamic regularizer and the first order condition for local optima become:
\begin{equation}
    \begin{split}
            R_{S_{k,j}}(\theta;  \theta_{S_{k,j}}^{t-1},  \theta_{S_{k,j-1}}^t) & \triangleq L_{S_{k,j}}(\theta) -\langle \nabla L_{S_{k},j}(\theta_{S_{k,j}}^{t-1}) \,, \theta \rangle + \frac{\alpha_{dyn}}{2} ||\theta -\theta_{S_{k,j-1}}^t||^2) \\
            \nabla R_{S_{k,j}}(\theta;  \theta_{S_{k,j}}^{t-1},  \theta_{S_{k,j-1}}^t) & = L_{S_{k},j}(\theta_{S_{k,j}}^{t}) -\nabla L_{S_{k},j}(\theta_{S_{k,j}}^{t-1}) + \alpha_{dyn} (\theta -\theta_{S_{k,j-1}}^t)
    \end{split}
\end{equation}
Applying the same reasoning of FedDyn, as $\theta_{S_{k,j}}^{t} \to \theta_{S_{k,j}}^{\infty}$ and $\nabla L_{S_{k},j}(\theta_{S_{k,j}}^{t}) \to \nabla L_{S_{k},j}(\theta_{S_{k,j}}^{\infty})$, this implies $\theta^t \to \theta_{S_{k,j}}^{\infty}$. Analogously the server side aggregation becomes:
\begin{equation}
    \theta^t= \frac{1}{|P_t|} \sum_{k \in P_t} \theta_{S_k}^t - \frac{1}{m} \sum_{k \in P_t}(\theta_{S_k}^t - \theta^{t-1})
\end{equation}
Because of the sequential training of the models, the term $\theta_{S_k}^t$ can be rewritten as the sum of the gradients computed by each client inside a superclient, leading to the following equation:
\begin{equation}
\begin{split}
    \theta_{S_k}^t &= \theta^{t-1} + \sum_j^{|S_k|} \nabla L_{S_{k,j}}(\theta_{S_{k,j}}^t) \Rightarrow \sum_{k \in P_t}(\theta_{S_k}^t - \theta^{t-1}) = \sum_{k \in P_t} \sum_j^{|S_k|} \nabla L_{S_{k,j}}(\theta_{S_{k,j}}^t) \\
    \theta^t &= \frac{1}{|P_t|} \sum_{k \in P_t} \theta_{S_k}^t - \frac{1}{m} \sum_{k \in P_t}\nabla L_{S_k}(\theta_{S_k}^t) \quad \textnormal{where} \quad \nabla L_{S_k}(\theta_{S_k}^t) \triangleq \sum_j^{|S_k|} \nabla L_{S_{k,j}}(\theta_{S_{k,j}}^t)
\end{split}
\end{equation}
In this way $\theta^t$ convergence implies $\sum_{k \in P_t}\nabla L_{S_k}(\theta_{S_k}^t) \to 0$: indeed the definitions of $h^t \triangleq \sum_k \nabla L_k(\theta_k^t)$ in FedDyn and $h^t \triangleq \sum_k \nabla L_{S_k}(\theta_{S_k}^t)$ in FedSeq are analogous.

\subsection{\textsc{Analysis on the superclients' local epochs $E_S$}}
\label{app:ES}
\begin{figure}[t]
\begin{center}
\includegraphics[width=\linewidth]{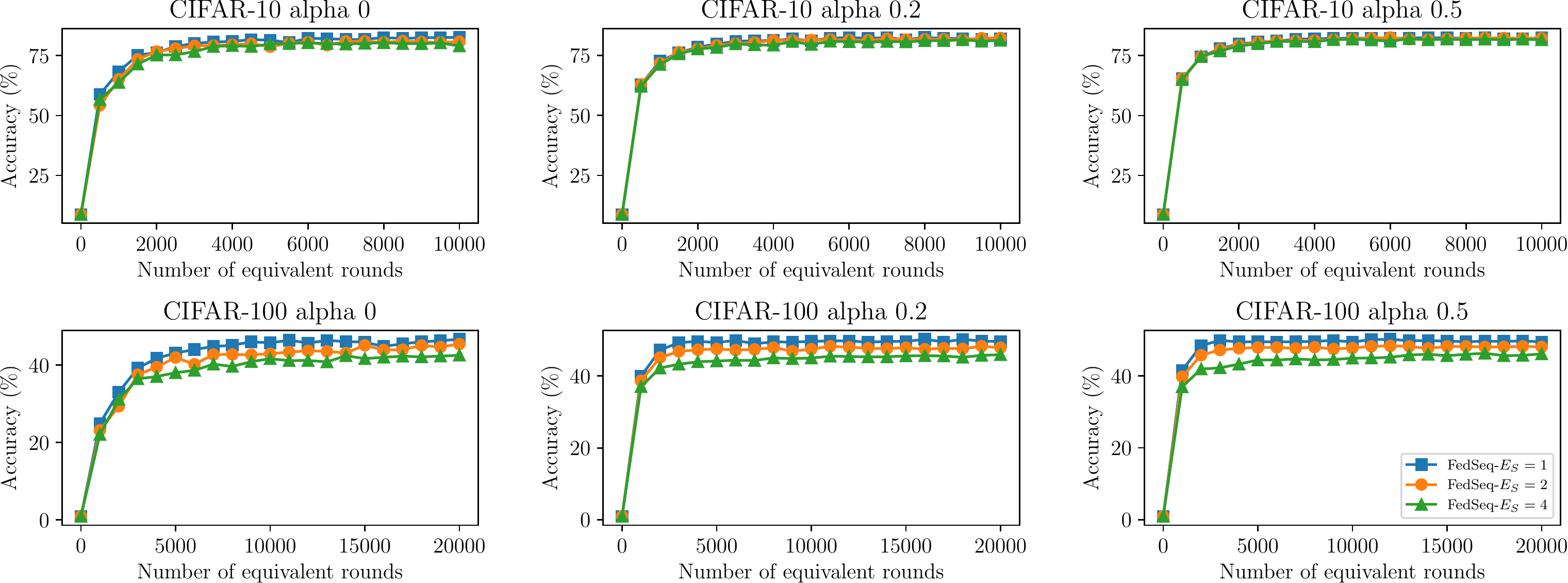}
\end{center}
   \caption{FedSeq varying $E_S \in \{1, 2, 4\}$. Results show that, on equal effort, increasing the amount of computation through superclients' clients does not improve the performance. Best viewed in color.}
\label{fig:ablationES}
\end{figure}

In analogy with the number of client's local epochs $E_k$, we analyse what happens increasing the superclient's epochs $E_S$. The intuition behind this study is that, since superclients are built on top of heterogeneous data distributions, more loops on their dataset could produce more robust models, not biased towards a single class. Increasing $E_S$ while decreasing the global round number $T$ does not impact the communication steps, but reduces the number of aggregations and accounts for a more loose synchronization. To compare fairly, Figure \ref{fig:ablationES} shows the results for $E_S \in \{1, 2, 4\}$, and $T$ decreased by the corresponding factor: to ease the visulization we compare the accuracies along \textit{equivalent rounds}, meaning that the actual round $r_{\textnormal{abs}}$ for each line in the graph is scaled by the number of $E_S$, formally $r_{\textnormal{eq}}=\frac{r_{\textnormal{abs}}}{E_S}$.
It is possible to notice that, comparing models with the same amount of training, increasing the number of sequential rounds among superclients' clients does not improve the performance accordingly. As increasing $E_S$ does not decrease communication cost, there is no advantage performing more than one epoch.
Differently, using the strategy of FedSeqInter, we obtain a similar effect in that models are more trained before the aggregation step, but:
\begin{itemize}
    \item \textbf{The dataset the model is sequentially trained on is broader:} indeed the aggregation period $N_S$ is choosen such that statistically the models encounters the whole global dataset before the aggregation step;
    \item \textbf{We do not add any computation per round:} even better, aggregation every $N_S$ rounds requires less sync.
\end{itemize}
In Section \ref{sec:experiments} of the main paper we empirically demonstrate that the latter approach ultimately leads to better convergence performances.


%% file: tables/ablation_cluster.tex
\begin{table*}[h]
    \begin{center}
    \caption{Comparison between clustering methods. Each result is the average of the scores obtained for $\alpha \in \{0, 0.2, 0.5\}$.}
    \label{tab:ablation_clusters}
    \resizebox{\linewidth}{!}{
    \begin{tabular}{l l l c c c c} 
        \toprule
        \multirow{2}{*}{Approximator $\psi$} & \multirow{2}{*}{Method $\phi$} & \multirow{2}{*}{Metric $\tau$} & \multicolumn{2}{c}{\textsc{Cifar-10}} & \multicolumn{2}{c}{\textsc{Cifar-100}}  \\ \cline{4-7}
        &  &  & Balance Ratio  & Covered Classes& Balance Ratio  & Covered Classes\\
        \midrule
        \multirow{3}{*}{classifierAll} & \multirow{2}{*}{Greedy} & Cosine distance & \textbf{0.334} & 0.886 & \textbf{0.028} & 0.667 \\
        & & Wasserstein distance & 0.081 & 0.759 & 0.011 & 0.651\\ 
        & K-means & Euclidean distance& 0.207 & 0.902 & 0.009 & 0.655 \\
        \midrule
        \multirow{3}{*}{classifierLast2} & \multirow{2}{*}{Greedy} & Cosine distance & \textbf{0.275} & 0.871 & \textbf{0.034} & 0.668 \\
        &  & Wasserstein distance & 0.090 & 0.746 & 0.009 & 0.652 \\ 
        & K-means & Euclidean distance & 0.203 & 0.900 & 0.009 & 0.654 \\
        \midrule
        \multirow{3}{*}{classifierLast} & \multirow{2}{*}{Greedy} & Cosine distance & \textbf{0.266} & 0.880 & \textbf{0.043} & 0.668\\
        & & Wasserstein distance & 0.085 & 0.755 & 0.010 & 0.650\\ 
        & K-means & Euclidean distance& 0.204 & 0.902 & 0.009 & 0.655\\
        \midrule
        \multirow{5}{*}{\shortstack[l]{confidence\\vectors}}      & \multirow{4}{*}{Greedy} & Cosine distance & \textbf{0.311} & 0.886 & \textbf{0.014} & 0.658\\
        & & Wasserstein distance & 0.077 & 0.784 & 0.009 & 0.654 \\
        & & KL divergence & 0.271 & 0.870 & 0.011 & 0.656\\
        & & Gini index & 0.298 & 0.876 & 0.012 & 0.657\\ 
        & K-means & Euclidean distance & 0.173 & 0.894 & 0.009 & 0.656\\ \midrule
        - & Random & - & 0.068 & 0.835 & 0.009 & 0.655\\
        \bottomrule
    \end{tabular}}
    \end{center}
\end{table*}